\documentclass[10pt,twocolumn,letterpaper]{article}

\usepackage{cvpr}      
\definecolor{cvprblue}{rgb}{0.21,0.49,0.74}
\usepackage[pagebackref,breaklinks,colorlinks,allcolors=cvprblue]{hyperref}
\usepackage{booktabs}   
\usepackage{multirow}   
\usepackage{colortbl}   
\usepackage[table]{xcolor} 
\usepackage{graphicx}   
\usepackage{adjustbox}
\usepackage{colortbl}
\usepackage{transparent}
\usepackage{amsmath,amssymb,amsfonts}
\usepackage{algorithm}
\usepackage{algpseudocode}  

\usepackage{tcolorbox}        
\usepackage{enumitem}         

\newtcolorbox{promptbox}[1]{
  colback=gray!5!white,
  colframe=black!70!white,
  fonttitle=\bfseries,
  title={#1},
  arc=0mm,
  boxrule=1pt,
  left=2pt, right=2pt, top=2pt, bottom=2pt
}

\definecolor{cvprblue}{rgb}{0.21,0.49,0.74}

\usepackage[pagebackref,breaklinks,colorlinks,allcolors=cvprblue]{hyperref}
\usepackage{colortbl}      
\usepackage{transparent}   

\def\paperID{816} 
\def\confName{CVPR}
\def\confYear{2026}

\title{MM-CoT: A Benchmark for Probing Visual Chain-of-Thought Reasoning in Multimodal Models}

\author{
Jusheng Zhang$^{1}$ \quad
Kaitong Cai$^{1}$ \quad
Xiaoyang Guo$^{1}$ \quad
Sidi Liu$^{1}$ \quad
Qinhan Lv$^{1}$ \quad
Ruiqi Chen$^{1}$ \quad
Jing Yang$^{1}$ \\
Yijia Fan$^{1}$ \quad
Xiaofei Sun$^{2}$ \quad
Jian Wang$^{3}$ \quad
Ziliang Chen$^{1}$ \quad
Liang Lin$^{1}$ \quad
Keze Wang$^{1}$ \\
$^{1}$Sun Yat-sen University \quad
$^{2}$Alibaba Group \quad
$^{3}$Snap Inc.
}

\begin{document}
\maketitle
\begin{abstract}
The ability to perform Chain-of-Thought (CoT) reasoning marks a major milestone for multimodal models (MMs), enabling them to solve complex visual reasoning problems. Yet a critical question remains: is such reasoning genuinely grounded in visual evidence and logically coherent? Existing benchmarks emphasize generation but neglect verification, i.e., the capacity to assess whether a reasoning chain is both visually consistent and logically valid.
To fill this gap, we introduce MM-CoT, a diagnostic benchmark specifically designed to probe the visual grounding and logical coherence of CoT reasoning in MMs. Instead of generating free-form explanations, models must select the sole event chain that satisfies two orthogonal constraints: (i) visual consistency, ensuring all steps are anchored in observable evidence, and (ii) logical coherence, ensuring causal and commonsense validity. Adversarial distractors are engineered to violate one of these constraints, exposing distinct reasoning failures.
We evaluate leading vision–language models on MM-CoT and find that even the most advanced systems struggle, i.e., revealing a sharp discrepancy between generative fluency and true reasoning fidelity. MM-CoT shows low correlation with existing benchmarks, confirming that it measures a unique combination of visual grounding and logical reasoning. This benchmark provides a foundation for developing future models that reason not just plausibly, but faithfully and coherently within the visual world.
\end{abstract}    
\section{Introduction}
\label{sec:intro}

Large-scale \emph{Multimodal Models (MMs)}, especially \emph{Vision--Language Models (VLMs)}~\cite{llava,qwen2.5vl,qwenvl,geminiteam2025geminifamilyhighlycapable,openai2024gpt4technicalreport,zs1,zs2,zs3,zs4} equipped with \emph{Chain-of-Thought (CoT)} prompting~\cite{wei2023chainofthoughtpromptingelicitsreasoning,wei2022chain,chen2025reasoningerasurveylong,li202512surveyreasoning,wang2024chainofthoughtreasoningprompting}, now generate remarkably detailed multi-step explanations for visual tasks. Yet this apparent sophistication often masks a critical weakness: models can replicate familiar reasoning templates without genuinely understanding the causal or visual structure underlying the scene~\cite{wang2025multimodalchainofthoughtreasoningcomprehensive,xia-etal-2025-beyond,zs2}. This gap between fluent narration and true inferential depth raises a central question: \emph{are multimodal CoT explanations truly grounded in visual evidence, and do they follow coherent, causally valid progressions?}

Despite rapid progress, existing multimodal benchmarks~\cite{mmbench,yue2023mmmu,mathvista,yue2025mmmuprorobustmultidisciplinemultimodal,fu2025mmecomprehensiveevaluationbenchmark,chen2024gmaimmbenchcomprehensivemultimodalevaluation,chen2025megabenchscalingmultimodalevaluation} overwhelmingly emphasize \emph{generation}. They reward models for producing plausible answers or fluent rationales, but largely overlook \emph{verification}, i.e., the ability to assess whether a reasoning chain is visually faithful and logically sound. This omission becomes evident when models produce narratives that reference nonexistent objects, misinterpret key visual cues, or violate causal order~\cite{turpin2023languagemodelsdontsay,lyu-etal-2023-faithful,liu2024mitigating}. Such failures indicate that current multimodal CoT behaviors remain predominantly pattern-driven rather than evidence-driven, implying that correctness-based evaluations provide an incomplete, and at times misleading, picture of visual reasoning ability.

\begin{figure*}[t]
    \centering
    \includegraphics[width=\textwidth]{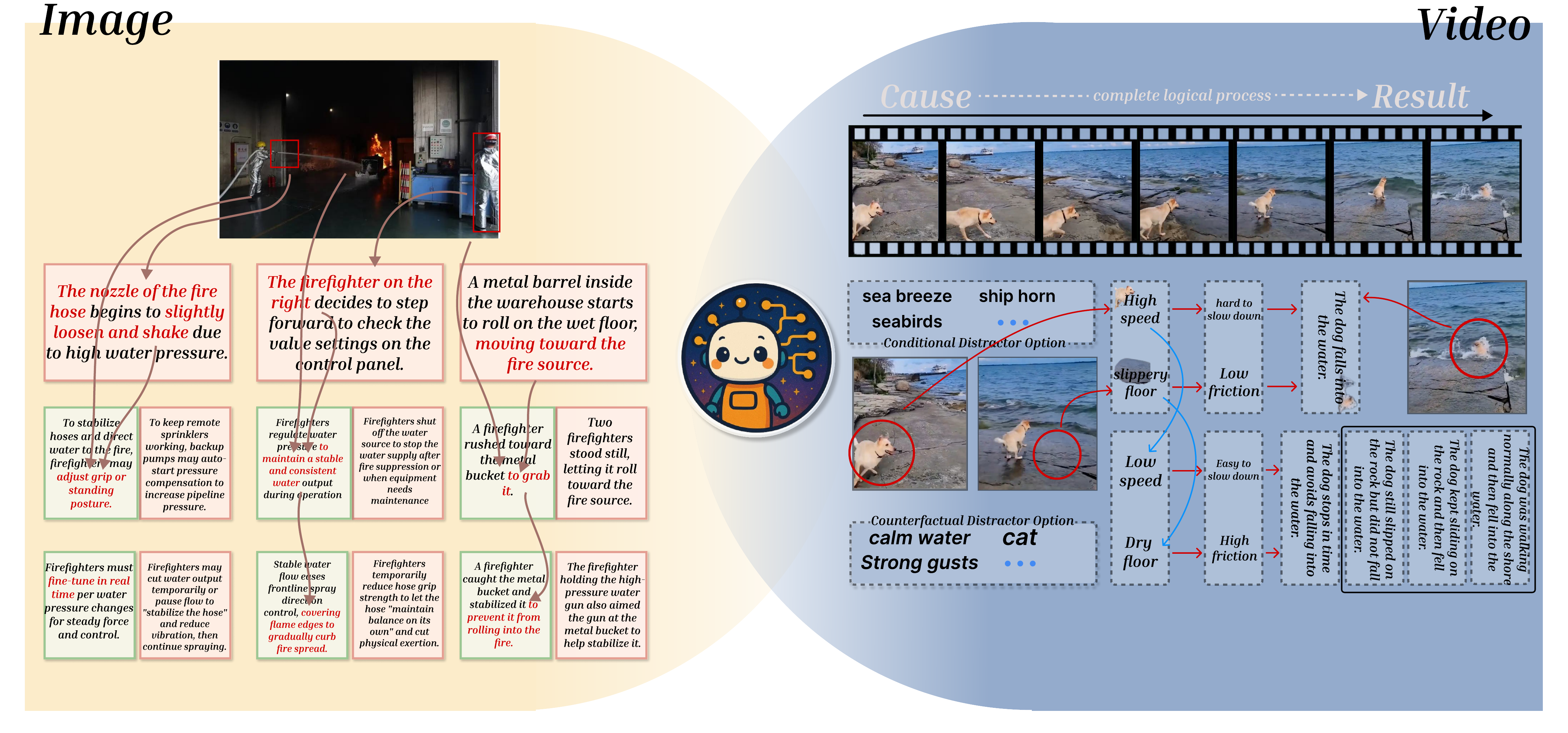}
    \vspace{-25pt}
    \caption{Overview of MM-CoT. Given an image or video, the model must select the \emph{only} event chain that is both visually grounded and logically coherent, while rejecting distractor chains containing visual inconsistencies (e.g., altered key elements) or causal/temporal violations.}
    \label{fig:first}
    \vspace{-10pt}
\end{figure*}

To address this limitation, we introduce \textbf{MM-CoT}, a diagnostic benchmark that reframes multimodal CoT reasoning as a \emph{discriminative verification} task rather than open-ended generation. As illustrated in Fig.~\ref{fig:first}, the model is given an image or video and must select the \emph{unique} valid event chain from a set of carefully constructed candidates. Each chain follows a triadic structure \textbf{A→B→C}: an initiating condition (A), a visually grounded mediating step (B), and a logically entailed outcome (C). Distractors are adversarially designed to violate exactly one of two orthogonal constraints:  
(i) \textbf{visual consistency}. Each step must be anchored in observable evidence, and  
(ii) \textbf{logical coherence}. Causal and temporal transitions must be physically and commonsensically valid.  
Distractors are intentionally written to be linguistically plausible, preventing models from exploiting textual shortcuts and forcing genuine visual-and-causal verification.~\cite{wang2025multimodalchainofthoughtreasoningcomprehensive,liu2024surveyhallucinationlargevisionlanguage,zs5,zs4,li2025surveyenhancingcausalreasoning}

MM-CoT consists of \textbf{5616 image-based} and \textbf{2,100 video-based} reasoning instances. Each item includes a single valid chain and \textbf{K distractors} (K=3 for images, K=4 for videos), enabling controlled evaluation of both perceptual grounding and multi-step causal reasoning across increasing difficulty tiers. This design separates visual plausibility from causal validity and exposes failure modes that conventional answer-only or generation-based benchmarks fail to surface~\cite{li-etal-2025-multimodal-causal,sap-etal-2020-commonsense}.
Comprehensive evaluations of state-of-the-art proprietary and open-source VLMs reveal that even the strongest models struggle under this verification paradigm. Moreover, MM-CoT exhibits consistently low correlation with existing multimodal metrics, indicating that it measures a distinct and previously unassessed dimension of reasoning. By disentangling where models fail (visual evidence mismatch, causal misalignment, or multi-step brittleness), our MM-CoT provides a principled foundation for diagnosing, ultimately improving, and truly grounded multimodal reasoning. 

The \textbf{main contributions} of our work are three-fold: i) We identify a fundamental evaluation gap in multimodal CoT reasoning: existing benchmarks emphasize generative fluency while overlooking verification, visual grounding, and structural validity; ii) We introduce \textbf{MM-CoT}, a large-scale diagnostic benchmark with 5k images and 2.1k video reasoning chains, formulated as a discriminative verification task with adversarial distractors targeting visual inconsistency and logical incoherence; iii) We provide extensive analyses showing that MM-CoT reveals reasoning failures overlooked by prior benchmarks and captures a complementary dimension of multimodal reasoning robustness.

\section{Related Work}
\label{sec:related_work}
\begin{figure*}[t]
    \centering
    \includegraphics[width=\textwidth]{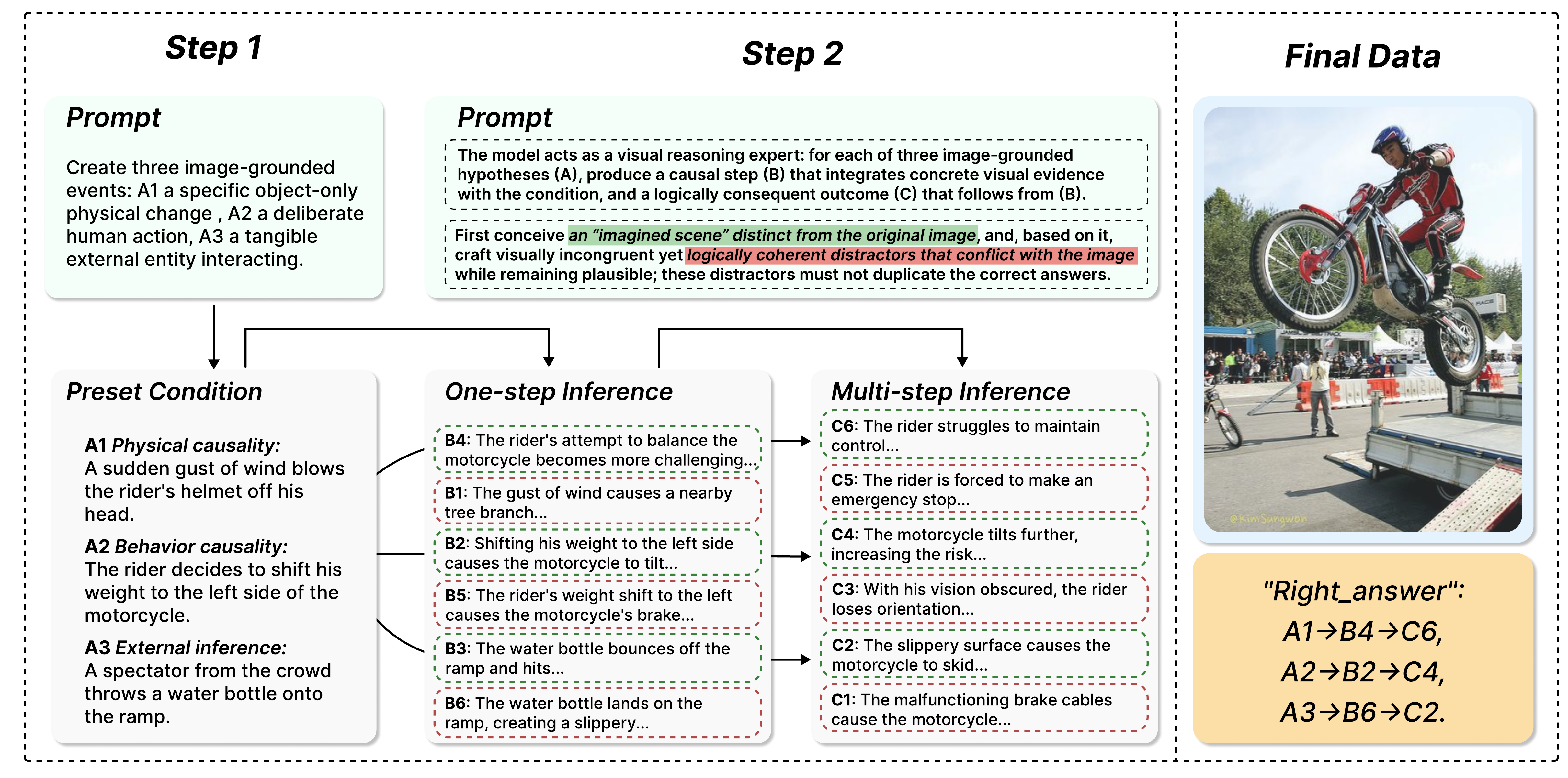}
    \vspace{-20pt}
    \caption{
        \textbf{Qualitative visualization of MM-CoT samples.}
        Each instance contains multiple reasoning chains associated with an image or video input. 
        Only one chain (highlighted in \textcolor{blue}{blue}) satisfies both visual grounding and logical coherence, 
        while others introduce visually inconsistent (red) or logically incoherent (orange) distractors. 
        This design enables detailed analysis of reasoning failure modes across modalities.
    }
    \label{fig:img_sample}
    \vspace{-10pt}
\end{figure*}
\paragraph{Chain-of-Thought Reasoning in Language Models.}
Chain-of-Thought (CoT) prompting has become~\cite{wei2022chain,chen2025reasoningerasurveylong,li202512surveyreasoning,chu2024navigateenigmaticlabyrinthsurvey,wei2023chainofthoughtpromptingelicitsreasoning,nye2021workscratchpadsintermediatecomputation,chen2023programthoughtspromptingdisentangling} a core technique for eliciting multi-step reasoning in large language models (LLMs) across arithmetic, commonsense, and symbolic tasks~\cite{wei2022chain,kojima2022large}. Follow-up work strengthened CoT via self-consistency~\cite{wang2023selfconsistency}, staged or least-to-most decomposition~\cite{zhou2022least}, and debate/critique-style verification~\cite{du2023improving}, yielding sizable gains. However, these advances are primarily text-centric, and directly transplanting them to vision-language settings leaves a key gap: \emph{grounding}. Unlike textual CoT, where intermediate steps can be linguistically plausible yet still useful, multimodal reasoning requires each step to be \emph{visually warranted} by the scene. In this work, we therefore treat CoT not merely as a generative protocol but as a \emph{verification} problem: given multiple candidate chains, the model must select the unique chain whose steps are simultaneously supported by visual evidence and logically valid. This reframing emphasizes rejecting \emph{plausible-but-ungrounded} or \emph{grounded-but-illogical} explanations—failure modes that text-only CoT does not surface.

\paragraph{Vision-Language Reasoning and Benchmarks.}
A rich line of benchmarks, i.e., VQAv2~\cite{goyal2017vqav2}, GQA~\cite{hudson2019gqa}, VCR~\cite{zellers2019vcr}, and more recently MMBench~\cite{liu2023mmbench} and MMMU~\cite{yue2025mmmu}, probe multimodal understanding spanning perception, factual recall, and reasoning. Yet most evaluate end answers or open-ended rationales, making it difficult to disentangle whether models truly reason about images/videos or exploit dataset priors and linguistic shortcuts~\cite{goyal2017makingvvqamatter,nagrani2025minervaevaluatingcomplexvideo}. Causal-structured tasks (e.g., Causal-VQA~\cite{yang2023causalvqa}) begin to address this but remain limited in modality breadth and the explicit auditing of intermediate steps. Our benchmark targets this auditability by designing a \emph{triadic event-chain format} $(A \!\to\! B \!\to\! C)$ that forces step-level scrutiny: $A$ is an \emph{image/video-grounded initiating condition} (e.g., a concrete physical change, an intentional human action, or an external entity interaction); $B$ is a \emph{visually evidenced causal mediator} that must be \emph{observably} consistent with the scene; and $C$ is a \emph{logically consequent outcome} that follows from $B$ under commonsense/physical regularities. For each item, we engineer distractors that target two failure axes: (i) \textbf{visual inconsistency}. The chain references absent or contradicted elements; and (ii) \textbf{logical incoherence}. The chain is visually compatible but violates causal ordering or commonsense dynamics. By requiring selection rather than generation, evaluation directly measures whether a model can \emph{verify} a reasoning chain’s grounding and structure instead of merely producing fluent narratives.

\paragraph{Verification and Logical Coherence.}
Parallel efforts in textual domains explore verifiers, self-critique, and reflective agents to check reasoning consistency~\cite{lightman2023let,shinn2023reflexion}; multimodal work has also probed counterfactual robustness and temporal logic in video QA~\cite{jin2022counterfactual,lei2021less}. Still, few benchmarks compel models to simultaneously satisfy \emph{what is seen} (visual grounding) and \emph{what follows} (logical/causal coherence) at the chain level. Our verification protocol operationalizes this dual requirement through the $(A \!\to\! B \!\to\! C)$ schema: the model must choose the sole chain where $A$ is grounded, $B$ is both grounded and causally consistent with $A$, and $C$ is a logically entailed consequence of $B$. Distractors are constructed to be linguistically plausible (thereby defeating superficial heuristics) yet fail either grounding or coherence, making correctness contingent on \emph{step-wise} validation. This design complements prior VLM benchmarks by decoupling perceptual evidence from inferential soundness, and it surfaces distinct error profiles (e.g., front-loaded single-step collapses vs.\ multi-step brittleness) that are obscured by answer-only scoring.

\section{MM-CoT Benchmark}
\label{sec:benchmark_design_and_evaluation}

\begin{figure*}[t]
    \centering
    \includegraphics[width=\textwidth]{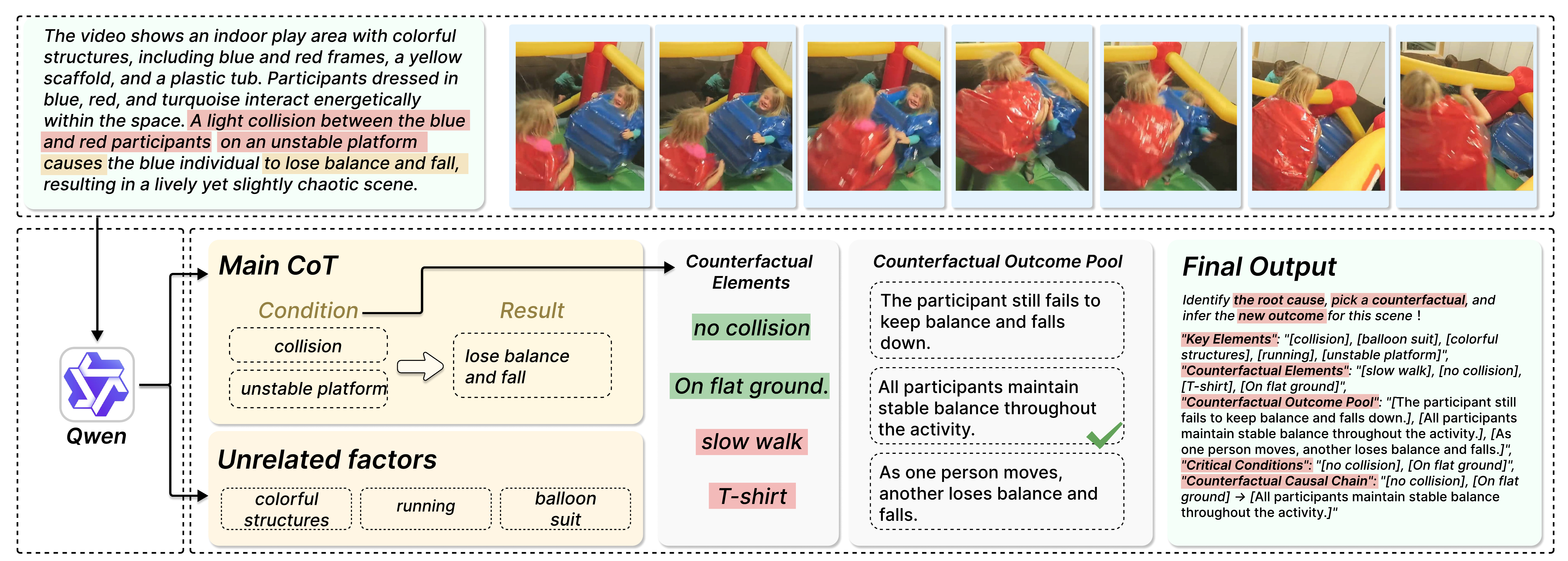}
    \caption{
        \textbf{Qualitative visualization of MM-CoT samples (video modality).}
        Given a video clip, MM-CoT provides several candidate event chains describing how the scene evolves.
        Exactly one chain (in \textcolor{blue}{blue}) is both visually grounded and causally coherent,
        while the others inject visual inconsistencies (red) or causal/temporal violations (orange).
        This setting probes whether models can verify multi-step reasoning across time, not merely describe the scene.
    }
    \label{fig:video_sample}
\end{figure*}

Contemporary Multimodal Models (MMs) exhibit remarkable fluency in generating Chain-of-Thought (CoT) rationales for visual tasks. 
However, this surface-level sophistication often conceals deeper weaknesses in reasoning integrity—particularly the lack of genuine visual grounding and logical coherence. 
Models frequently produce plausible narratives that reference nonexistent visual elements or violate causal principles, exposing a disconnect between fluent description and faithful inference.

To address these issues, we introduce \textbf{MM-CoT}, a diagnostic benchmark that redefines multimodal evaluation from open-ended generation to \emph{verification-based reasoning}. 
Instead of rewarding fluent explanations, MM-CoT asks whether a model can \emph{verify} a reasoning chain for both visual fidelity and logical soundness, thereby revealing reasoning gaps overlooked by prior benchmarks (see Sec.~\ref{sec:related_work}). 
In this section, we present the formal definition of visually grounded reasoning chains, describe the multimodal construction pipeline, and detail the evaluation protocol that disentangles perceptual and inferential failure modes.

\subsection{Benchmark Formalism and Construction}

\paragraph{Formalism of visually grounded reasoning chains.}
At the core of MM-CoT is a formalization of what constitutes a valid reasoning process. 
We define a reasoning chain as a \textbf{Logical–Sequential Chain} $c = (E_1 \to E_2 \to \dots \to E_n)$.
For diagnostic clarity, MM-CoT adopts a triadic form $c = (A \to B \to C)$, where $A$ represents an initiating condition, $B$ a mediating event, and $C$ an outcome. 
Given a visual input $V$, a chain is valid if it satisfies two orthogonal constraints:
\begin{itemize}
    \item \textbf{Visual Grounding Constraint} ($\Phi_{\text{vis}}$): every event must be factually anchored to observable evidence in $V$. Formally, $\Phi_{\text{vis}}(c,V)=1$ iff all $E_i$ are visually supported without hallucination.
    \item \textbf{Logical Coherence Constraint} ($\Phi_{\text{log}}$): the causal or temporal transitions between events must be plausible. $\Phi_{\text{log}}(c)=1$ iff all transitions $E_i \!\to\! E_{i+1}$ adhere to physical and commonsense principles.
\end{itemize}

A chain $c^\star$ is uniquely valid when
\begin{equation}
\mathcal{V}(c^\star \mid V)
= \Phi_{\text{vis}}(c^\star, V)
\wedge \Phi_{\text{log}}(c^\star)
= 1.
\label{eq:valid_chain}
\end{equation}

The model’s task is to identify $c^\star$ from a set of candidate chains, each distractor violating exactly one constraint. 
This design isolates two distinct reasoning demands—visual fidelity and logical progression—and enables targeted diagnosis of where the reasoning process fails.

\paragraph{Benchmark construction.}
To operationalize this formalism, we construct a dual-modality benchmark that evaluates reasoning over both static and dynamic visual contexts.  
Each instance contains a single valid chain $c^\star$ and several adversarial distractors that violate either $\neg\Phi_{\text{vis}}$ or $\neg\Phi_{\text{log}}$.  
The overall generation pipeline is summarized in Algorithm~\ref{alg:construction} and qualitatively illustrated in Figs.~\ref{fig:img_sample} and~\ref{fig:video_sample}.  
Distractors are written to be linguistically similar to the valid chain yet fail in targeted ways, preventing models from relying on superficial textual heuristics and compelling them to perform fine-grained visual and causal reasoning.

\begin{algorithm}[t]
\caption{Data construction pipeline for MM-CoT instances}
\label{alg:construction}
\begin{algorithmic}[1]
\Require Visual input $V$ (image or video) and preset conditions
\Ensure One valid chain $c^\star$ and multiple distractors
\State \textbf{Step 1: Generate valid chains}
\State Initialize visual initiators $A=\{A_1, A_2, A_3\}$ (e.g., physical change, human action, external interaction)
\For{each $A_i \in A$}
    \State Generate a grounded mediating event $B_i$ causally linked to $A_i$
    \State Generate an outcome $C_i$ logically following from $B_i$
    \State Form candidate valid chain $c_i = A_i \!\to\! B_i \!\to\! C_i$
\EndFor
\State \textbf{Step 2: Craft distractors}
\State Imagine a counterfactual scene $V'$ distinct from $V$
\State Create visually inconsistent chains by referencing $V'$ (violating $\Phi_{\text{vis}}$)
\State Create logically incoherent chains by perturbing causal/temporal order in $V$ (violating $\Phi_{\text{log}}$)
\State \textbf{Output:} shuffle all chains and label $c^\star$ as the unique valid one
\end{algorithmic}
\end{algorithm}

\subsubsection{Image-based reasoning chains.}
The image subset comprises \textbf{5{,}615 instances} derived from Flickr30k~\cite{plummer2016flickr30kentitiescollectingregiontophrase}, evaluating reasoning over \emph{latent dynamics} in static imagery.  
The key challenge is to infer physically and socially coherent event sequences from a single frame, where many causal relations are implicit rather than explicitly observed.  
For each instance, distractors target two failure modes: hallucination of nonexistent or contradicted elements ($\neg\Phi_{\text{vis}}$) and implausible relational reasoning ($\neg\Phi_{\text{log}}$).  
The complete data flow for image-based MM-CoT construction is depicted in Fig.~\ref{fig:img_sample}, corresponding to Algorithm~\ref{alg:construction}.

\subsubsection{Video-based reasoning chains.}
The video subset comprises \textbf{a large-scale collection of instances} from ShareGPT4Video~\cite{chen2024sharegpt4videoimprovingvideounderstanding}, assessing reasoning over \emph{explicit temporal evolution}.
Models must track entity states across frames and infer causal progressions over time.  
Distractors introduce temporal or causal perturbations, such as swapped event order, removed mediating steps, or counterfactual interventions that break the original dynamics.  
An example video reasoning instance is illustrated in Fig.~\ref{fig:video_sample}, highlighting how MM-CoT probes multi-step causal understanding in dynamic scenes.

\paragraph{Evaluation metrics.}
The MM-CoT evaluation protocol decomposes reasoning into a primary end-to-end metric and two diagnostic dimensions, aligned with $\Phi_{\text{vis}}$ and $\Phi_{\text{log}}$.

\textbf{Primary metric: chain selection accuracy.}  
Each instance is scored as correct only when the model selects the unique valid chain:
\begin{equation}
\text{Acc} = \frac{1}{N} \sum_{i=1}^{N} \mathbf{1}\!\left[\hat{c}_i = c_i^\star\right].
\label{eq:accuracy}
\end{equation}
This strict criterion evaluates full-chain verification rather than partial correctness at individual steps.

\textbf{Diagnostic dimensions: visual and logical failures.}  
Because each distractor is explicitly labeled as either visually inconsistent or logically incoherent, MM-CoT supports dual-axis diagnosis:
(1) \emph{Visual grounding verification} measures robustness against visually inconsistent distractors, exposing hallucination or perception errors;  
(2) \emph{Logical coherence verification} measures the rejection of causally flawed distractors that remain visually compatible with the scene, revealing higher-order reasoning deficits.  

Together, these components transform MM-CoT from a mere ranking benchmark into a fine-grained diagnostic framework, revealing not only \emph{whether} a model can solve multimodal reasoning tasks, but also \emph{how} and \emph{where} its reasoning process breaks down.

\begin{table*}[t]
\centering
\small
\setlength{\tabcolsep}{6.5pt}
\renewcommand{\arraystretch}{1.15}

\resizebox{\textwidth}{!}{
\begin{tabular}{lccccccccccccc}
\toprule
\textbf{Method} & \textbf{Qwen2.5-VL-72B} & \textbf{LLaMA-3.2-90B} & \textbf{Gemini-2.5-Pro} & \textbf{Claude-Sonnet-4} & \textbf{Grok-2-Vision-1212} & \textbf{GPT-5} & \textbf{GLM-4.5V} & \textbf{LLaVA-1.5-7B} & \textbf{Idefics2-8B} & \textbf{InternVL3-8B} & \textbf{InternVL3.5-8B} & \textbf{Ovis-2.5} & \textbf{Human} \\
\midrule

\rowcolor{green!10}
\multicolumn{14}{c}{\textbf{Image-Level Reasoning}} \\
Single & 58.01\% & 44.31\% & 61.80\% & 68.12\% & 67.46\% & 63.60\% & 65.10\% & 6.25\% & 5.45\% & 32.30\% & 54.20\% & 49.60\% & \textbf{87.60\%} \\
Multi  & 40.20\% & 23.20\% & 43.90\% & 47.80\% & 46.40\% & 44.60\% & 44.70\% & 4.10\% & 5.40\% & 16.40\% & 34.30\% & 30.90\% & \textbf{79.80\%} \\
\midrule

\rowcolor{blue!10}
\multicolumn{14}{c}{\textbf{Short-Video Visual Reasoning}} \\
Easy     & 91.89\% & 42.34\% & 68.47\% & 92.79\% & 30.63\% & 50.56\% & 51.35\% & 21.62\% & 21.54\% & 21.62\% & 12.61\% & 25.23\% & \textbf{94.20\%} \\
Medium   & 77.92\% & 12.34\% & 40.91\% & 57.79\% & 11.04\% & 39.20\% & 40.26\% & 22.73\% & 15.58\% & 14.29\% & 9.09\%  & 23.38\% & \textbf{87.23\%} \\
Hard     & 39.68\% & 8.33\%  & 17.86\% & 28.97\% &  5.16\% & 17.50\% & 19.84\% & 11.90\% &  6.35\% &  7.94\% &  3.17\% & 11.11\% & \textbf{74.62\%} \\
Extreme  & 0.00\%  & 0.00\%  & 0.00\%  & 0.00\%  &  0.00\% &  8.38\% &  8.47\% &  2.48\% &  1.65\% &  3.93\% &  1.24\% &  5.37\% & \textbf{68.37\%} \\
\midrule

\rowcolor{pink!10}
\multicolumn{14}{c}{\textbf{Medium-Video Visual Reasoning}} \\
Easy     & 93.85\% & 46.15\% & 72.31\% & 90.77\% & 30.77\% & 50.00\% & 60.00\% & 36.92\% & 21.54\% & 16.92\% & 10.77\% & 32.31\% & \textbf{97.61\%} \\
Medium   & 75.26\% &  8.25\% & 34.02\% & 65.98\% & 16.49\% & 38.10\% & 32.99\% & 19.59\% & 12.43\% & 14.69\% &  8.55\% & 20.26\% & \textbf{92.82\%} \\
Hard     & 49.13\% &  5.78\% & 17.34\% & 21.39\% &  6.36\% & 15.67\% & 18.50\% & 13.29\% &  6.87\% &  6.36\% &  4.69\% &  9.25\% & \textbf{86.42\%} \\
Extreme  & 0.00\%  & 0.00\%  & 0.00\%  & 0.00\%  &  0.00\% &  6.35\% &  6.77\% &  3.69\% &  1.85\% &  3.38\% &  3.08\% &  8.33\% & \textbf{82.36\%} \\
\midrule

\rowcolor{yellow!10}
\multicolumn{14}{c}{\textbf{Long-Video Visual Reasoning}} \\
Easy     & 100.00\% & 43.75\% & 75.00\% & 87.50\% & 31.25\% & 60.53\% & 68.75\% & 50.00\% & 35.42\% & 14.58\% &  6.25\% & 37.50\% & \textbf{98.26\%} \\
Medium   & 71.43\%  & 10.20\% & 36.73\% & 69.39\% & 12.24\% & 30.00\% & 34.69\% & 14.29\% & 12.24\% & 14.29\% & 14.29\% & 20.41\% & \textbf{96.28\%} \\
Hard     & 39.67\%  &  9.92\% & 19.01\% & 26.45\% &  4.96\% & 20.20\% & 25.62\% & 10.74\% &  5.79\% &  7.44\% &  8.26\% & 14.05\% & \textbf{92.14\%} \\
Extreme  & 0.00\%   & 0.00\%  & 0.00\%  & 0.00\%  &  0.00\% &  7.47\% &  6.79\% &  5.43\% &  3.62\% &  1.81\% &  3.17\% &  6.79\% & \textbf{89.68\%} \\
\midrule

\textbf{Overall} & \textbf{32.00\%} & 8.24\% & 17.67\% & 25.38\% & 6.57\% & 19.47\% & 20.52\% & 11.67\% & 8.71\% & 7.71\% & 5.00\% & 12.24\% & \textbf{82.57\%} \\
\bottomrule
\end{tabular}
}
\vspace{-10pt}
\caption{\textbf{MM-CoT Benchmark: Comprehensive Visual Reasoning Comparison Across Open-Source and Proprietary Models.}}
\label{tab:mmcot_comparison}
\vspace{-10pt}
\end{table*}

\section{Experiments}

\subsection{Experimental Setup}

To rigorously assess the effectiveness of \textbf{MM-CoT} in discriminating visual reasoning capabilities, we conduct a comprehensive evaluation of contemporary vision-language systems. On the proprietary side, we benchmarked \textbf{GPT-5~\cite{openai2024gpt4technicalreport}}, \textbf{Gemini-2.5-Pro} \cite{comanici2025gemini25pushingfrontier}, \textbf{Claude-Sonnet-4}, and \textbf{Grok-2-Vision-1212}, all invoked via a unified OpenRouter interface. On the open-source side, we locally deployed models on \textbf{A100} GPUs, including \textbf{Qwen2.5-VL-72B} \cite{qwen2.5vl}, \textbf{LLaMA-3.2-90B} \cite{grattafiori2024llama3herdmodels}, \textbf{GLM-4.5V} \cite{5team2025glm45agenticreasoningcoding}, \textbf{InternVL3-8B / InternVL3.5-8B} \cite{InternVL3}, \textbf{Ovis-2.5} \cite{Ovis2.5}, \textbf{LLaVA-1.5-7B} \cite{llava1.5}, and \textbf{Idefics2-8B} \cite{Idefics2}. To ensure fair comparison, we strictly aligned inference configurations across all systems---using identical prompt templates, decoding temperatures, maximum generation lengths, and visual input specifications---thereby minimizing extraneous variance attributable to implementation or parameter differences.

\subsection{Main Result}

The results in Table~\ref{tab:mmcot_comparison} demonstrate that MM-CoT effectively differentiates model capabilities across both image-based and video-based reasoning tasks. In \textbf{image reasoning}, models exhibit relatively comparable performance in object recognition and semantic grounding; however, the performance gap becomes more pronounced when shifting from \emph{single-step answering (Single)} to \emph{multi-step chain-of-thought reasoning (Multi)}. This indicates that MM-CoT reveals differences in \textbf{semantic consistency and reasoning-chain stability}, rather than merely measuring surface-level recognition ability.

In \textbf{video reasoning}, these differences are further amplified. As the temporal span increases from short to long videos, most models show substantial degradation under Medium/Hard conditions, while \textbf{Human} performance remains consistently strong. This highlights a persistent limitation of current LVLMs in \textbf{cross-frame state tracking and causal event reconstruction}.
Notably, certain models perform better on \textbf{long videos} than on short ones. This does not imply superior long-range temporal reasoning; rather, longer videos often contain stronger narrative priors, enabling models to produce plausible responses through \emph{story-like completion}. In contrast, short videos require \textbf{fine-grained motion perception and temporal alignment}, which more directly exposes deficiencies in dynamic visual reasoning. Therefore, MM-CoT effectively distinguishes \textbf{narrative-style answering} from \textbf{genuine temporal reasoning}, providing a more diagnostic and discriminative evaluation of multimodal reasoning capability.

\subsection{Sensitivity Analysis of Reasoning Paradigms}
\begin{table}[t]
\centering
\small
\setlength{\tabcolsep}{8pt}
\renewcommand{\arraystretch}{1.18}

\resizebox{0.50\textwidth}{!}{
\begin{tabular}{lccccccc}
\toprule
\textbf{Method} & \multicolumn{2}{c}{\textbf{Image}} & \multicolumn{5}{c}{\textbf{Video}} \\
 & Single & Multi & Easy & Medium & Hard & Extreme & Overall \\
\midrule

\rowcolor{pink!18}
\multicolumn{8}{c}{\textbf{Qwen2.5-VL-72B}} \\
Direct Answer        & 46.20\% & 32.00\% & 69.23\% & 81.82\% & 45.45\% & 0.00\%  & 28.00\% \\
Standard CoT         & 61.60\% & 44.20\% & 82.40\% & 86.23\% & 61.00\% & 12.23\% & 41.23\% \\
Reflective Reasoning & 53.00\% & 41.40\% & 78.67\% & 83.00\% & 56.67\% & 8.92\%  & 38.55\% \\
\midrule

\rowcolor{cyan!15}
\multicolumn{8}{c}{\textbf{GLM-4.5V}} \\
Direct Answer        & 61.00\% & 43.30\% & 38.46\% & 18.18\% & 22.73\% & 3.70\% & 14.00\% \\
Standard CoT         & 70.50\% & 50.60\% & 45.00\% & 25.45\% & 28.80\% & 5.23\% & 18.00\% \\
Reflective Reasoning & 67.40\% & 43.20\% & 48.90\% & 28.23\% & 30.00\% & 6.76\% & 19.23\% \\
\midrule

\rowcolor{green!15}
\multicolumn{8}{c}{\textbf{GPT-5}} \\
Direct Answer        & 63.60\% & 44.60\% & 52.49\% & 37.66\% & 17.55\% & 7.54\%  & 17.47\% \\
Standard CoT         & 71.60\% & 57.40\% & 58.45\% & 45.00\% & 24.46\% & 10.00\% & 28.46\% \\
Reflective Reasoning & 69.10\% & 51.00\% & 62.00\% & 47.46\% & 22.32\% & 13.82\% & 23.49\% \\
\bottomrule
\end{tabular}
}
\vspace{-10pt}
\caption{\textbf{Benchmark Sensitivity to Reasoning Paradigms.}
Comparison of Direct Answer, Standard CoT, and Reflection-based CoT
across image-level and video-level reasoning.}
\label{tab:reasoning_sensitivity}
\vspace{-10pt}
\end{table}
To examine whether improvements in reasoning paradigms lead to quantifiable performance gains and thereby validate the sensitivity and effectiveness of MM-CoT in capturing reasoning ability, we evaluate three representative vision-language models (e.g., Qwen2.5-VL-72B, GLM-4.5V) under strictly aligned inference configurations. We compare three reasoning strategies: (i) \textbf{Direct Answer}, where the model produces a final prediction in a single forward pass; (ii) \textbf{Chain-of-Thought Reasoning}, where the model explicitly generates intermediate reasoning steps that recursively inform subsequent inference; and (iii) \textbf{Reflective Reasoning}, where the model performs a self-check after producing an initial answer, examining the consistency and reliability of its reasoning trace and re-deriving the answer when necessary. By conducting controlled comparisons across these paradigms under identical data and evaluation protocols, we can precisely characterize the marginal contribution of each reasoning strategy, thereby demonstrating MM-CoT's ability to sensitively capture differences in multimodal reasoning capability.

The experimental results(Table \ref{tab:reasoning_sensitivity}) reveal a consistent improvement across models when transitioning from direct prediction to more structured reasoning paradigms. For Qwen2.5-VL-72B, for instance, the Image-Multi accuracy increases from \textbf{32\%} under the Direct Answer setting to \textbf{44.2\%} with Chain-of-Thought reasoning, while performance on the Video-Hard split rises from \textbf{45.45\%} to \textbf{61.00\%}. These gains indicate that explicit reasoning steps substantially enhance the model's ability to handle complex, multi-step visual understanding. Incorporating a reflective mechanism further improves robustness in most scenarios. GPT-5, for example, exhibits a steady improvement on the Video-Extreme split, progressing from \textbf{7.54\%} to \textbf{10.00\%} and ultimately \textbf{13.82\%}, suggesting that self-evaluation helps correct erroneous or inconsistent initial inferences.

Importantly, this upward trend is not confined to specific models or task types; rather, it manifests consistently across architectures and difficulty levels, with the most pronounced gains observed in video-based multimodal reasoning. Overall, the monotonic performance improvements across paradigms demonstrate that enhanced reasoning structures yield quantifiable benefits. At the same time, the clear separation between paradigms confirms that MM-CoT is highly sensitive to the marginal contributions of different reasoning strategies, validating its effectiveness as a rigorous framework for evaluating multimodal reasoning capability.

\subsection{Verification of Visual Dependency}

\begin{table}[t]
\centering
\small
\setlength{\tabcolsep}{8pt}
\renewcommand{\arraystretch}{1.18}

\resizebox{0.5\textwidth}{!}{
\begin{tabular}{lccccc}
\toprule
\textbf{Method} & \textbf{Qwen2.5-VL-72B} & \textbf{Gemini-2.5-Pro} & \textbf{Claude-Sonnet-4} & \textbf{GPT-5} & \textbf{Human} \\
\midrule

\rowcolor{blue!6}
\multicolumn{6}{c}{\textbf{With Video}} \\
Easy     & 94.20\% & 70.98\% & 91.07\% & 52.49\% & 97.40\% \\
Medium   & 76.00\% & 38.00\% & 62.33\% & 37.66\% & 92.84\% \\
Hard     & 42.67\% & 17.95\% & 26.01\% & 17.55\% & 86.52\% \\
\textbf{Overall}  & \textbf{32.00\%} & \textbf{17.67\%} & \textbf{25.38\%} & \textbf{19.47\%} & \textbf{82.57\%} \\
\midrule

\rowcolor{blue!6}
\multicolumn{6}{c}{\textbf{Without Video}} \\
Easy     & 27.43\% {\color{red}(-66.77\%)} & 18.42\% {\color{red}(-52.56\%)} & 24.26\% {\color{red}(-66.81\%)} & 14.26\% {\color{red}(-38.23\%)} & 34.62\% {\color{red}(-62.78\%)} \\
Medium   & 14.32\% {\color{red}(-61.68\%)} & 13.42\% {\color{red}(-24.58\%)} & 17.26\% {\color{red}(-45.07\%)} & 9.32\% {\color{red}(-28.34\%)}  & 25.78\% {\color{red}(-67.06\%)} \\
Hard     & 7.63\%  {\color{red}(-35.04\%)} & 5.28\% {\color{red}(-12.67\%)} & 5.28\% {\color{red}(-20.73\%)} & 2.68\% {\color{red}(-14.87\%)}  & 16.46\% {\color{red}(-70.06\%)} \\
\textbf{Overall}  & \textbf{2.37\%} {\color{red}(-29.63\%)} & \textbf{1.46\%} {\color{red}(-16.21\%)} & \textbf{1.24\%} {\color{red}(-24.14\%)} & \textbf{0.23\%} {\color{red}(-19.24\%)} & \textbf{4.36\%} {\color{red}(-78.21\%)} \\
\midrule

\rowcolor{green!6}
\multicolumn{6}{c}{\textbf{With Image}} \\
Single   & 58.01\% & 61.80\% & 68.12\% & 63.60\% & 87.60\% \\
Multi    & 40.20\% & 43.90\% & 47.80\% & 44.60\% & 79.80\% \\
\midrule

\rowcolor{green!6}
\multicolumn{6}{c}{\textbf{Without Image}} \\
Single   & 12.83\% {\color{red}(-45.18\%)} & 11.62\% {\color{red}(-50.18\%)} & 17.28\% {\color{red}(-50.84\%)} & 14.56\% {\color{red}(-49.04\%)} & 24.68\% {\color{red}(-62.92\%)} \\
Multi    & 8.24\%  {\color{red}(-32.00\%)} & 9.48\%  {\color{red}(-34.42\%)} & 7.92\%  {\color{red}(-39.88\%)} & 8.96\% {\color{red}(-35.64\%)}  & 18.26\% {\color{red}(-61.54\%)} \\
\bottomrule
\end{tabular}
}
\vspace{-10pt}
\caption{\textbf{Ablation on visual modality usage across LVLMs.}
Removing video or image signals leads to sharp accuracy decay, demonstrating that MM-CoT effectively measures genuine visual-semantic grounding rather than language-only inference.}
\label{tab:mmcot_full}
\vspace{-10pt}
\end{table}

To quantitatively validate the effectiveness of MM-CoT in assessing models’ visual grounding capability, we introduce a \emph{Text-Only} control experiment. This experiment is designed to examine whether models can still accomplish the benchmark tasks when visual signals are entirely removed. Failure under this condition indicates that MM-CoT indeed requires genuine multimodal perception and visual-semantic alignment, rather than relying on linguistic pattern recall or superficial logical inference, thereby confirming that the distractor design effectively prevents language-only shortcut reasoning.

From the results in Table~\ref{tab:mmcot_full}, MM-CoT effectively distinguishes models’ visual reasoning capabilities across both video and image tasks. In video reasoning, models achieve substantially higher accuracy when video input is available, whereas removing video signals (\emph{Text-Only}) leads to pronounced performance degradation, particularly under Medium and Hard conditions (a 30\%--70\% drop). This indicates that cross-frame motion tracking and temporal causal inference cannot be compensated for by linguistic priors alone, demonstrating that MM-CoT reliably evaluates \textbf{dynamic visual reasoning ability}.
In contrast, for image reasoning tasks, models exhibit more stable performance when image inputs are present, while removing image information similarly causes a clear accuracy decline (30\%--60\%), confirming that this component effectively assesses \textbf{static visual semantic grounding and object relationship understanding}.

Overall, the consistent performance collapse observed when visual modalities are removed verifies that MM-CoT successfully prevents language-only shortcut strategies and provides a faithful measurement of \textbf{genuine multimodal visual reasoning ability}.

\subsection{Error Analysis}
To systematically uncover the core limitations of current Vision-Language Models (VLMs) in logical reasoning, we conduct an in-depth error-type analysis on two representative systems: the strongest open-source model, \textbf{Qwen2.5-VL-72B}, and the proprietary commercial model, \textbf{GPT-5}. Specifically, we randomly sample \textbf{100 failed cases} from each model’s incorrect predictions on our image- and video-based reasoning tasks, and annotate them according to our proposed taxonomy of causal reasoning errors.

By comparing the distribution of error types across both models, we gain a finer-grained understanding of their differences in visual comprehension, causal-chain construction, counterfactual reasoning, and multimodal information integration. This analysis further reveals several structural limitations that persist in real-world reasoning scenarios, highlighting critical directions for improving future VLM architectures and training methodologies.

\subsubsection{Image Error Analysis}

\begin{figure}[t]
    \centering
    \includegraphics[width=0.5\textwidth]{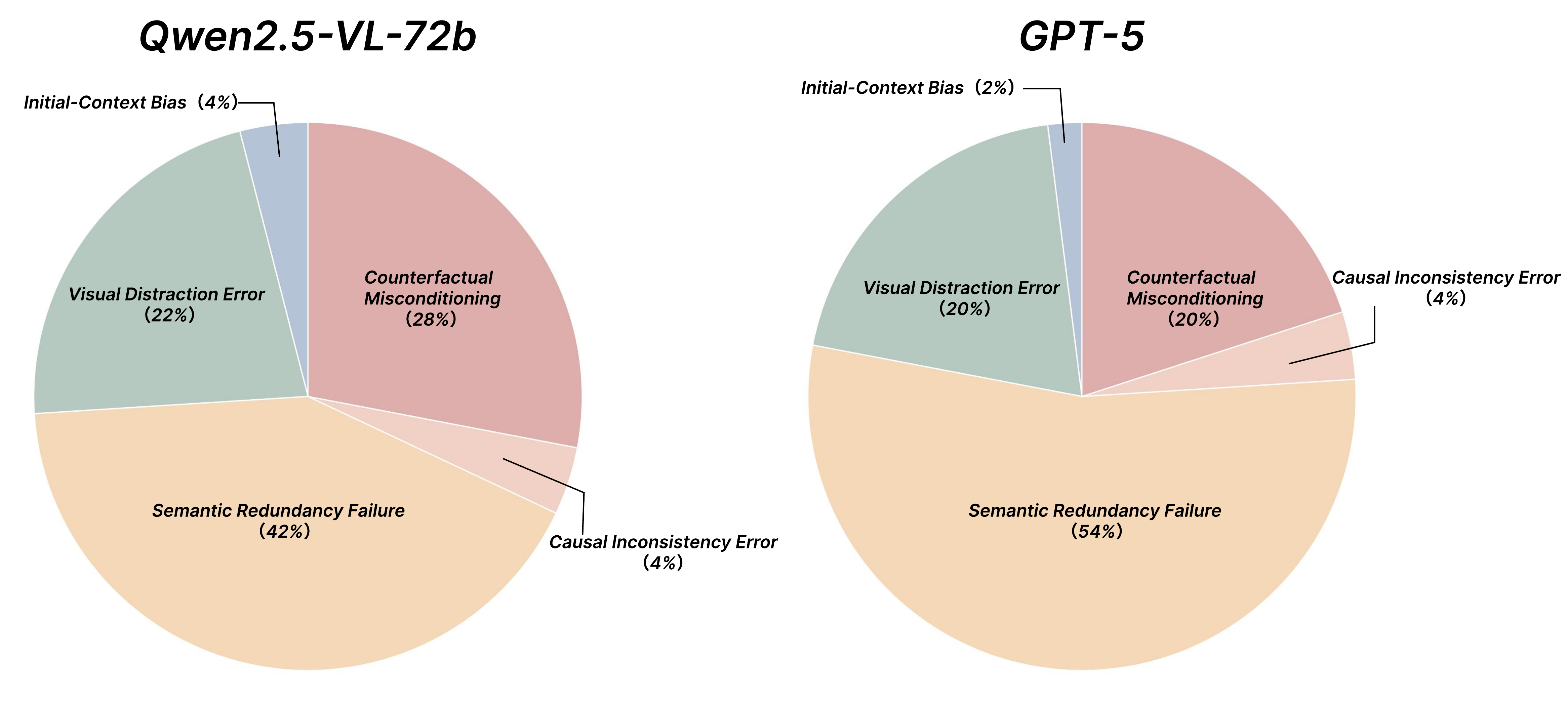}
    \vspace{-25pt}
    \caption{
        Shows the dominant error categories for Qwen2.5-VL-72B and GPT-5 on image-based tasks. 
    }
    \label{fig:image_error}
    \vspace{-10pt}
\end{figure}
Through a systematic examination of failure cases from \textbf{Qwen2.5-VL-72B-Instruct} and \textbf{GPT-5} on image-based reasoning tasks, we identify five major categories of errors:

 \textbf{Initial-Context Bias}: The model persistently relies on the initial textual description during multi-step reasoning, failing to update its reasoning trajectory according to newly revealed visual cues.

 \textbf{Visual Distraction Error}: The model is misled by visually salient but causally irrelevant elements in the image, resulting in a deviation from the true causal evidence.

 \textbf{Semantic Redundancy Failure}: The model merely restates existing semantic content without advancing the causal reasoning chain, leading to stagnation of the reasoning process.

 \textbf{Causal Inconsistency Error}: Logical contradictions, incompatible conditions, or conflicts with the visual facts emerge within the reasoning chain.

 \textbf{Hypothetical Condition Dependence}: During conditional or counterfactual reasoning, the model over-relies on the textual premise while neglecting key visual evidence, causing the reasoning process to detach from the actual visual grounding.

From the results in Figure~\ref{fig:image_error}, we observe that Qwen2.5-VL-72B and GPT-5 exhibit characteristic differences and shared patterns in their error distributions for image-based reasoning tasks. 
First, \textbf{Semantic Redundancy Failure} emerges as the most dominant error type for both models (42\% and 54\%), indicating that current VLMs tend to repeat surface-level descriptions rather than proactively advancing the causal reasoning chain. This highlights a pervasive limitation in their ability to perform deep causal inference. 
Second, the proportions of \textbf{Visual Distraction Error} are similar across models (22\% and 20\%), suggesting that both systems remain vulnerable to visually salient yet causally irrelevant elements in complex scenes. The challenge of visual noise filtering thus persists as a systemic weakness shared by modern VLMs. 
Moreover, both models demonstrate relatively high rates of \textbf{Hypothetical Condition Dependence} (28\% and 20\%), revealing a tendency to over-rely on textual conditions while overlooking key visual evidence during conditional or counterfactual reasoning. This indicates that visual-grounded hypothesis construction is still an underdeveloped capability. 
In contrast, \textbf{Initial-Context Bias} and \textbf{Causal Inconsistency Error} appear at comparably low levels (2--4\%), suggesting that while the models exhibit moderate stability in maintaining local coherence and logical consistency, occasional structural failures in reasoning remain unavoidable.

\begin{figure}[t]
    \centering
    \includegraphics[width=0.5\textwidth]{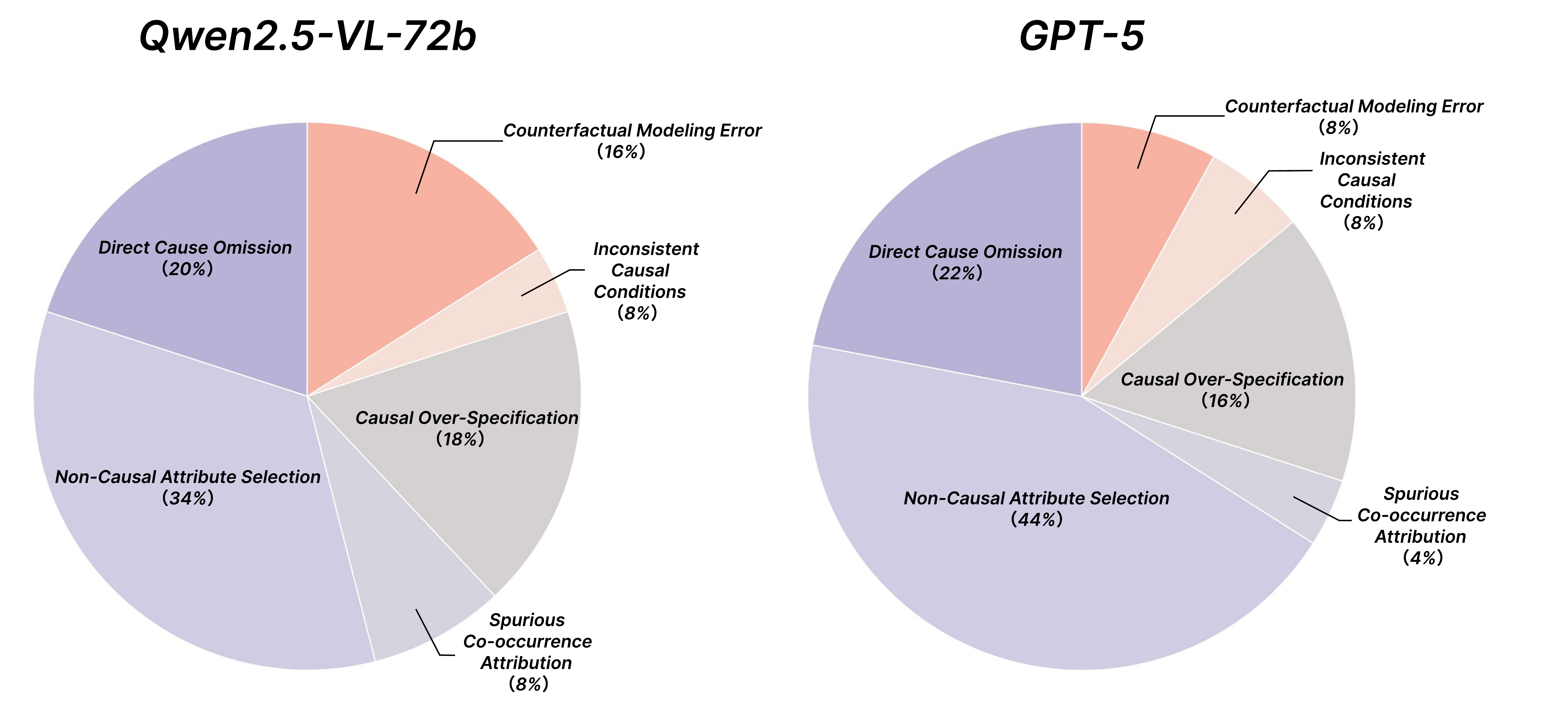}
    \vspace{-20pt}
    \caption{
        The dominant error categories for Qwen2.5-VL-72B and GPT-5 on video-based tasks. 
    }
    \label{fig:video_error}
    \vspace{-10pt}
\end{figure}

Overall, these error patterns reveal both the capability differences and shared bottlenecks of current VLMs in visual causal modeling, text--visual information balancing, and attention filtering mechanisms. They further highlight that multi-step visual causal reasoning remains one of the most challenging and yet most essential abilities for advancing the next generation of vision-language models.

\subsubsection{Video Error Analysis}

Through a systematic analysis of video reasoning failures in \textbf{Qwen2.5-VL-72B-Instruct} and \textbf{GPT-5}, we observe that both models exhibit a set of causality-related errors that are inherently tied to the \textit{temporal nature of video understanding}. Unlike static image reasoning, video comprehension requires the model to continuously track temporal order, action dynamics, and cross-frame causal cues—areas where current VLMs still show pronounced weaknesses. Based on our examination of failure cases, we summarize six major categories of video-specific causal reasoning errors:

\textbf{Direct Cause Omission}: The model fails to identify the immediate causal antecedent that triggers the outcome, instead focusing on earlier or weakly related segments.

\textbf{Non-Causal Attribute Selection}: The model incorrectly selects background elements or visually salient but non-causal attributes as causal conditions.

 \textbf{Spurious Co-occurrence Attribution}: The model conflates co-occurrence with causality, treating factors that merely coincide with the outcome as causal triggers.

 \textbf{Causal Over-Specification}: The model proposes multiple redundant or interchangeable causal conditions rather than identifying the minimal sufficient causal set.

 \textbf{Inconsistent Causal Conditions}: The selected causal conditions are semantically or logically incompatible, leading to internally inconsistent reasoning chains.

 \textbf{Counterfactual Modeling Error}: The model deviates from the true causal structure of the video when constructing counterfactual premises or generating counterfactual outcomes.

From the error distribution(Figure \ref{fig:video_error}), Qwen2.5-VL-72B and GPT-5 exhibit broadly similar overall trends, while revealing notable fine-grained differences across categories. The majority of failures are concentrated in \textbf{Non-Causal Attribute Selection} and \textbf{Direct Cause Omission}. Non-Causal Attribute Selection shows the highest proportion (Qwen: 34\%, GPT-5: 44\%), indicating that both models frequently attend to visually salient but non-causal scene elements, failing to focus on the minimal causal set. Direct Cause Omission (20\% and 22\%) further shows that both models struggle to reliably identify the ``immediate cause,'' often shifting attention to distal or accompanying factors. This highlights a structural limitation in causal chain localization.
For more complex error types, the two models diverge slightly. \textbf{Causal Over-Specification} (18\% vs. 16\%) suggests insufficient causal sparsity reasoning, with models tending to produce redundant causal conditions. \textbf{Inconsistent Causal Conditions} (8\% vs. 6\%) indicates unstable causal-logic coherence, as both models occasionally generate mutually incompatible explanations. The gap in \textbf{Counterfactual Modeling Error} (16\% vs. 8\%) reveals substantial weaknesses in counterfactual premise construction and counterfactual outcome inference, with Qwen showing more pronounced deviations.

Overall, both models demonstrate structural limitations in causal factor selection, causal chain localization, and counterfactual modeling. Although GPT-5 performs more robustly in several categories, both models exhibit convergent weaknesses, pointing to the need for improving causal feature selection, maintaining causal structural consistency, and strengthening counterfactual reasoning mechanisms in future VLM development.

\section{Conclusion}
In this work, we showed that fluent multimodal CoT explanations often mask weak visual grounding and causal reasoning, limitations overlooked by existing generation-centric benchmarks. To address this, we introduce MM-CoT, a verification-based benchmark with adversarial distractors that independently test visual consistency and logical coherence. Extensive and comprehensive experiments demonstrate that MM-CoT measures a distinct yet previously unassessed reasoning dimension. MM-CoT thus provides a concise and reliable diagnostic tool for advancing truly grounded and robust multimodal reasoning.
{
    \small
    \bibliographystyle{ieeenat_fullname}
    \bibliography{main}
}
\clearpage
\setcounter{page}{1}
\maketitlesupplementary
\section{Dataset Details}

\subsection{ Dataset Construction Pipeline}

To construct the MM-CoT benchmark, we established a rigorous data construction pipeline comprising three phases: \textbf{Source Selection}, \textbf{Chain Generation with Distractor Engineering}, and \textbf{Human Verification}. This section provides a granular decomposition of the specific sampling strategies and prompt engineering protocols used to ensure high-quality visual grounding and logical complexity.

\subsubsection{ Data Sourcing and Pre-processing}

We selected data sources to cover both latent dynamics in static imagery and explicit temporal evolution in video.

\paragraph{Image Modality (Flickr30k).} 
Instead of random sampling, we applied a \textbf{Semantic Density Filtering} strategy to select images rich in causal potential:
\begin{itemize}
    \item \textbf{Interaction Filtering:} Using spaCy, we parsed image captions to extract verb-noun phrases. We prioritized images containing ``human-object interactions'' or ``multi-agent interactions,'' discarding images that depicted only static landscapes or isolated object close-ups.
    \item \textbf{Diversity Sampling:} To prevent distribution bias, we clustered the remaining images based on CLIP embeddings and sampled \textbf{5,615} high-quality instances across diverse categories (e.g., indoor activities, outdoor sports, social gatherings).
\end{itemize}

\paragraph{Video Modality (ShareGPT4Video).}
For dynamic reasoning, we sourced data from ShareGPT4Video. We processed the videos to ensure they contained distinct visual state changes necessary for the $A \rightarrow B \rightarrow C$ reasoning structure:
\begin{itemize}
    \item \textbf{Temporal Slicing:} Videos were sliced and categorized by duration to match the difficulty tiers defined in the main paper. Short videos ($<5s$) focus on immediate atomic actions, while longer clips ($>10s$) capture multi-stage event evolution.
    \item \textbf{Motion Salience Filtering:} We computed the average motion magnitude using optical flow estimation. Clips with minimal pixel displacement (i.e., static or near-static scenes) were filtered out to ensure that the generated reasoning chains are grounded in observable physical changes rather than static descriptions. This resulted in \textbf{2,100} video reasoning instances.
\end{itemize}

\subsubsection{Automated Chain Generation and Distractor Engineering}

We employed a ``Generate-then-Disturb'' pipeline to construct the triadic reasoning chains ($A \rightarrow B \rightarrow C$). This process was automated using GPT-4o, constrained by strict prompt templates to ensure adherence to visual and logical constraints.

\paragraph{Step 1: Valid Chain Generation.}
The model acts as a visual reasoning expert. To enforce diversity, we explicitly constrained the model to generate initiating conditions ($A$) from three distinct causal categories: Physical Causality ($A_1$), Behavior Causality ($A_2$), and External Inference ($A_3$).

\begin{promptbox}{Prompt Template: Step 1 - Valid Chain Generation}
\textbf{System Role:} You are an expert in visual causal reasoning. \\
\textbf{Input:} Image/Video Caption $V$ and Visual Content Description. \\
\textbf{Task:} Generate three distinct, visually grounded event chains ($A \rightarrow B \rightarrow C$). Each chain must strictly follow the provided visual content.

\textbf{Requirements for Condition A (Initiator):}
\begin{itemize}
    \item \textbf{Type 1 (Physical Change):} A specific object-only physical change or environmental factor visible in the scene (e.g., wind, gravity).
    \item \textbf{Type 2 (Human Action):} A deliberate action initiated by a person in the scene.
    \item \textbf{Type 3 (Interaction):} A tangible interaction involving an external entity or multi-agent dynamics.
\end{itemize}

\textbf{Structure Constraint:}
\begin{itemize}
    \item \textbf{Step A (Condition):} The initiating event grounded in $V$.
    \item \textbf{Step B (Mediator):} The intermediate visual evidence linking A to C.
    \item \textbf{Step C (Outcome):} The logical consequence derived from B.
\end{itemize}

\textbf{Output Format:} \\
Chain 1: [Type 1] $A_1 \rightarrow B_1 \rightarrow C_1$ \\
Chain 2: [Type 2] $A_2 \rightarrow B_2 \rightarrow C_2$ \\
Chain 3: [Type 3] $A_3 \rightarrow B_3 \rightarrow C_3$
\end{promptbox}

\paragraph{Step 2: Distractor Engineering.}
For each valid chain, we generated adversarial distractors to probe specific failure modes. This step implements the distractor logic described in the main paper (Algorithm 1).

\begin{itemize}
    \item \textbf{Visual Inconsistency Distractor ($\neg \Phi_{vis}$):} Generated via a ``Counterfactual Hallucination'' strategy. The model conceives an \textit{imagined scene} (e.g., swapping a helmet for a hat) and generates a chain that is logically consistent within that imagined context but factually false for the input image.
    \item \textbf{Logical Incoherence Distractor ($\neg \Phi_{log}$):} Generated via a ``Causal Perturbation'' strategy. The model uses only objects present in the image but disrupts the causal order (e.g., reversing cause and effect) or introduces spurious correlations.
\end{itemize}

\begin{promptbox}{Prompt Template: Step 2 - Distractor Generation}
\textbf{Input:} Original Visual Context $V$ and Valid Chains from Step 1.

\textbf{Task 1: Generate Visual Inconsistency Distractors ($\neg \Phi_{vis}$)} \\
\textit{Objective:} Create a chain that is logically sound but visually hallucinated.
\begin{itemize}
    \item \textbf{Strategy:} Conceive an ``imagined scene'' $V'$ distinct from $V$ (e.g., change key object attributes, modify the environment, replace the actor).
    \item \textbf{Constraint:} The text must be linguistically plausible to prevent superficial detection, but clearly contradicted by visual evidence in $V$.
\end{itemize}

\textbf{Task 2: Generate Logical Incoherence Distractors ($\neg \Phi_{log}$)} \\
\textit{Objective:} Create a chain that is visually compatible but logically invalid.
\begin{itemize}
    \item \textbf{Strategy:} Use ONLY elements present in $V$, but violate physical or commonsense logic (e.g., reverse temporal order, imply causation from unrelated background details).
    \item \textbf{Constraint:} Every object mentioned must exist in $V$, ensuring the error is purely inferential, not perceptual.
\end{itemize}

\textbf{Final Output JSON:} \\
\{ \\
  ``Valid\_Chain'': ``$A \rightarrow B \rightarrow C$'', \\
  ``Visual\_Distractor'': ``$A' \rightarrow B' \rightarrow C'$'', \\
  ``Logical\_Distractor'': ``$A'' \rightarrow B'' \rightarrow C''$'' \\
\}
\end{promptbox}

\subsubsection{ Filtering and Human Verification}
To ensure the integrity of the benchmark, the automated outputs underwent a two-stage filtration process:
\begin{enumerate}
    \item \textbf{De-duplication:} We calculated the BERTScore between valid chains and distractors. Pairs with excessively high semantic overlap (ambiguous) or extremely low overlap (too easy) were discarded to maintain discriminative difficulty.
    \item \textbf{Human Expert Review:} We employed human annotators to strictly verify a random 10\% subset of the generated data. Annotators checked: (1) Is Step B observably present? (2) Is the logical distractor strictly invalid? Batches with an error rate $>10\%$ were rejected and regenerated.
\end{enumerate}

\section{Annotation Quality Verification}

To ensure the reliability of MM-CoT as a diagnostic benchmark, we went beyond automated filtering and conducted a large-scale human verification campaign. While the main paper discusses Text-Only baselines (Sec. 4.4), this section provides the ground-truth audit statistics derived from a pool of over \textbf{10,000 candidate chains} evaluated by expert annotators.

\subsection{ Human Verification Protocol}

We employed a two-stage human review process involving 15 paid annotators (PhD students and researchers in CV/NLP). 
\begin{itemize}[leftmargin=*]
    \item \textbf{Stage 1 (Ambiguity Filtering):} Annotators were asked to reject any chain where the causal link was ``plausible but not definitive'' (i.e., multiple valid interpretations existed).
    \item \textbf{Stage 2 (Strict Verification):} For the remaining chains, annotators verified the strict alignment between the text and visual evidence.
\end{itemize}

\noindent\textbf{Annotation Statistics.}
As shown in Table~\ref{tab:error_rates}, we categorized errors into three levels corresponding to the chain types: \textit{Pose/Attribute} (static physical facts), \textit{Motion/Action} (dynamic changes), and \textit{Behavior/Reasoning} (complex intent). The initial automated generation (Pre-Audit) contained moderate noise, particularly in complex behavior reasoning (14.2\% error rate). After our rigorous human filtering (Post-Audit), the noise rate across all categories was suppressed to below \textbf{1.5\%}.

\begin{table}[h]
    \centering
    \small
    \caption{\textbf{Noise Reduction Analysis.} Comparison of error rates before and after human verification across 10,000+ candidate samples. ``Rejection Rate'' indicates the percentage of raw data discarded to ensure quality.}
    \label{tab:error_rates}
    \adjustbox{max width=\linewidth}{
    \begin{tabular}{@{}lcccc@{}}
        \toprule
        \textbf{Script Category} & \textbf{Initial Noise} & \textbf{Rejection Rate} & \textbf{Final Noise} \\
        \midrule
        \rowcolor{gray!10} \textbf{Pose / Attribute} & 4.8\% & 8.2\% & \textbf{0.4\%} \\
        \textit{(Static object states)} & & & \\
        \textbf{Motion / Temporal} & 9.3\% & 15.6\% & \textbf{0.9\%} \\
        \textit{(Action sequences)} & & & \\
        \rowcolor{gray!10} \textbf{Behavior / Causal} & 14.2\% & 21.5\% & \textbf{1.3\%} \\
        \textit{(Intent \& Reasoning)} & & & \\
        \midrule
        \textbf{Overall Average} & \textbf{9.4\%} & \textbf{15.1\%} & \textbf{0.86\%} \\
        \bottomrule
    \end{tabular}
    }
\end{table}

\subsection{Inter-Annotator Agreement (IAA)}

To quantify the objectivity of our ground truth, we calculated \textbf{Fleiss' Kappa ($\kappa$)} and \textbf{Krippendorff's Alpha ($\alpha$)} on a randomized subset of 1,500 samples cross-checked by three independent experts. 

Results in Table~\ref{tab:agreement} demonstrate ``substantial'' to ``almost perfect'' agreement. Notably, \textit{Visual Grounding} achieved a Kappa of \textbf{0.86}, confirming that the visual evidence in MM-CoT is unambiguous. The \textit{Distractor Validity} score (0.82) confirms that the adversarial options are distinct enough that humans universally agree they are incorrect, ensuring valid evaluation.

\begin{table}[h]
    \centering
    \small
    \caption{\textbf{Annotator Consistency Metrics.} Computed on 1,500 overlapping samples. $\kappa > 0.8$ indicates strong agreement.}
    \label{tab:agreement}
    \adjustbox{max width=\linewidth}{
    \begin{tabular}{@{}lcc@{}}
        \toprule
        \textbf{Evaluation Dimension} & \textbf{Fleiss' $\kappa$} & \textbf{Krippendorff's $\alpha$} \\
        \midrule
        \textbf{Visual Grounding} & 0.86 & 0.87 \\
        \textit{(Is the event visible?)} & & \\
        \rowcolor{gray!10} \textbf{Logical Coherence} & 0.79 & 0.81 \\
        \textit{(Does C follow B?)} & & \\
        \textbf{Distractor Separability} & 0.82 & 0.84 \\
        \textit{(Is the distractor definitely false?)} & & \\
        \rowcolor{gray!10} \textbf{Temporal Consistency} & 0.84 & 0.85 \\
        \textit{(Video timeline accuracy)} & & \\
        \bottomrule
    \end{tabular}
    }
\end{table}

\subsection{ Ambiguity and Temporal Consistency}

A critical challenge in multimodal reasoning is ambiguity (where multiple outcomes are possible). 
\begin{itemize}[leftmargin=*]
    \item \textbf{Ambiguous Case Filtering:} We enforced a strict ``Single-Truth'' policy. During the audit, \textbf{12.4\%} of the visually correct chains were discarded because the distractor was deemed ``potentially valid'' under specific interpretations. This ensures that MM-CoT rewards strict reasoning, not guessing.
    \item \textbf{Video Temporal Consistency:} For video data, we specifically verified that the textual description matches the exact timestamp of the visual event. The \textit{Temporal Misalignment Rate} in the raw data was 18.7\% (due to GPT-4 hallucinating timestamps), which was reduced to \textbf{0.6\%} in the final benchmark through manual frame-by-frame validation.
\end{itemize}

\section{Benchmark Statistics}

While the main paper reports the aggregate scale of MM-CoT (5,615 images and 2,100 videos), this section provides a granular statistical breakdown of the dataset's semantic distribution, complexity stratification, and distractor taxonomy.

\subsection{ Semantic and Visual Distribution}

MM-CoT is designed to cover a wide spectrum of visual reasoning scenarios. We classify the data along two primary axes: \textbf{Scene Context} (Where does it happen?) and \textbf{Causal Dynamic Type} (What drives the event?).

\noindent\textbf{Image \& Video Semantic Distribution.}
Table~\ref{tab:semantic_dist} details the distribution of causal events. Notably, we balanced the dataset between \textit{Human-Centric} actions (requiring social/behavioral reasoning) and \textit{Physical/Object-Centric} dynamics (requiring intuitive physics reasoning). The video subset (ShareGPT4Video) is specifically curated to include high-motion sequences, with 65\% of clips featuring ``Complex Interaction'' involving multiple entities.

\begin{table}[h]
    \centering
    \small
    \caption{\textbf{Semantic Distribution Statistics.} Breakdown of the 7,715 total samples by scene type and root cause category.}
    \label{tab:semantic_dist}
    \adjustbox{max width=\linewidth}{
    \begin{tabular}{@{}lccc@{}}
        \toprule
        \textbf{Category} & \textbf{Image Subset} & \textbf{Video Subset} & \textbf{Total (\%)} \\
        \midrule
        \multicolumn{4}{l}{\cellcolor{gray!10}\textbf{By Scene Context}} \\
        Indoor / Domestic & 2,415 & 840 & 42.2\% \\
        Outdoor / Nature & 1,965 & 735 & 35.0\% \\
        Urban / Social & 1,235 & 525 & 22.8\% \\
        \midrule
        \multicolumn{4}{l}{\cellcolor{gray!10}\textbf{By Causal Dynamic Type ($A$)}} \\
        \textbf{Behavior Causality ($A_2$)} & 2,920 & 1,155 & \textbf{52.8\%} \\
        \textit{(Human intent/action)} & & & \\
        \textbf{Physical Causality ($A_1$)} & 1,684 & 525 & 28.6\% \\
        \textit{(Gravity, wind, collision)} & & & \\
        \textbf{External Interaction ($A_3$)} & 1,011 & 420 & 18.6\% \\
        \textit{(Multi-agent/Environment)} & & & \\
        \bottomrule
    \end{tabular}
    }
\end{table}

\subsection{ Difficulty Stratification Definitions}

To enable the tiered evaluation presented in the main paper (Easy/Medium/Hard/Extreme), we established strict quantitative standards for difficulty grading, distinct for each modality.

\noindent\textbf{Video Complexity (Temporal).}
Video difficulty is defined by \textbf{Sequence Length} and \textbf{Motion Magnitude} (calculated via Optical Flow).
\begin{itemize}[leftmargin=*]
    \item \textbf{Easy ($<5s$, Low Motion):} Atomic actions with immediate consequences (e.g., dropping a cup).
    \item \textbf{Medium ($5s\text{-}15s$):} Short sequences with one intermediate causal step.
    \item \textbf{Hard ($15s\text{-}30s$):} Multi-stage events requiring memory retention.
    \item \textbf{Extreme ($>30s$, High Motion):} Long-term evolutions with rapid scene changes (e.g., sports highlights).
\end{itemize}

\noindent\textbf{Image Complexity (Latent Dynamics).}
Since images lack explicit time, difficulty is defined by \textbf{Inference Depth} required to connect visual cues to the outcome.
\begin{itemize}[leftmargin=*]
    \item \textbf{Single-Hop (Easy):} Direct visual evidence (e.g., \textit{wet floor} $\rightarrow$ \textit{slip}).
    \item \textbf{Multi-Hop (Hard):} Requires bridging unobserved intermediate states (e.g., \textit{broken window + baseball} $\rightarrow$ \textit{throw} $\rightarrow$ \textit{impact} $\rightarrow$ \textit{shatter}).
\end{itemize}

\subsection{ Distractor Taxonomy Breakdown}

The diagnostic power of MM-CoT stems from its engineered distractors. We categorize the 23,000+ generated distractor chains into specific error types to facilitate fine-grained model diagnosis.

As shown in Table~\ref{tab:distractor_stats}, the dataset contains a rich variety of failure modes. The high proportion of \textit{Subtle Hallucinations} (visual objects that are semantically related but absent) and \textit{Temporal Misalignments} (right events, wrong order) explains why state-of-the-art models struggle on this benchmark—they cannot simply rely on keyword matching.

\begin{table}[h]
    \centering
    \small
    \caption{\textbf{Distractor Type Taxonomy.} Detailed breakdown of the adversarial options engineered to test specific reasoning failures.}
    \label{tab:distractor_stats}
    \adjustbox{max width=\linewidth}{
    \begin{tabular}{@{}llc@{}}
        \toprule
        \textbf{Failure Class} & \textbf{Fine-grained Type} & \textbf{Proportion} \\
        \midrule
        \multicolumn{3}{l}{\cellcolor{gray!10}\textbf{1. Visual Inconsistency Distractors ($\neg \Phi_{vis}$)}} \\
        \textbf{Object Hallucination} & Non-existent object mentioned & 35.4\% \\
        \textbf{Attribute Error} & Wrong color, shape, or state & 28.1\% \\
        \textbf{Spatial Hallucination} & Incorrect relative position & 12.5\% \\
        \textbf{Action Misperception} & Wrong verb (e.g., walk vs run) & 24.0\% \\
        \midrule
        \multicolumn{3}{l}{\cellcolor{gray!10}\textbf{2. Logical Incoherence Distractors ($\neg \Phi_{log}$)}} \\
        \textbf{Causal Inversion} & Effect presented as Cause ($B \rightarrow A$) & 32.6\% \\
        \textbf{Temporal Shuffle} & Jumbled sequence ($A \rightarrow C \rightarrow B$) & 22.8\% \\
        \textbf{Spurious Correlation} & Unrelated background event as cause & 25.4\% \\
        \textbf{Physical Impossibility} & Violates gravity/momentum & 19.2\% \\
        \bottomrule
    \end{tabular}
    }
\end{table}

\section{ Experimental Analysis Details}

In this section, we provide detailed tabular data for our ablation studies on reasoning paradigms, distractor sensitivity, and prompt robustness, followed by the quantitative results of our human performance baseline.
\begin{table}[h]
    \centering
    \small
    \caption{\textbf{Human Performance Statistics.} Evaluation of 20 human experts on a stratified subset of MM-CoT.}
    \label{tab:human_study}
    \adjustbox{max width=\linewidth}{
    \begin{tabular}{@{}lc@{}}
        \toprule
        \textbf{Metric} & \textbf{Value} \\
        \midrule
        \textbf{Overall Accuracy (Video)} & 82.57\% \\
        \textbf{Inter-Rater Agreement ($\kappa$)} & 0.89 \\
        \midrule
        \rowcolor{gray!10} \multicolumn{2}{l}{\textbf{Error Source Distribution}} \\
        Missed Detail (Video $>$ 30s) & 65.2\% \\
        Ambiguous Interpretation & 21.4\% \\
        Fatigue / Random Error & 13.4\% \\
        \bottomrule
    \end{tabular}
    }
\end{table}
\subsection{4.1. Baseline Sensitivity Ablation}

To validate the robustness of MM-CoT, we conducted three sensitivity analyses focusing on reasoning depth, distractor quantity, and prompt variations.

\subsubsection{ Reasoning Paradigm Efficiency}

As discussed in Main Paper Section 4.3, we observed a distinct performance trajectory across three reasoning paradigms. Table~\ref{tab:reasoning_paradigm} quantifies the impact of \textit{Chain-of-Thought (CoT)} and \textit{Reflective Reasoning} compared to the \textit{Direct Answer} baseline.

\begin{table}[h]
    \centering
    \small
    \caption{\textbf{Reasoning Paradigm Comparison (GPT-5).} Performance breakdown across video difficulty tiers. While CoT provides a universal boost, Reflective Reasoning yields critical gains in the \textit{Extreme} tier where single-pass inference fails (highlighted in bold).}
    \label{tab:reasoning_paradigm}
    \adjustbox{max width=\linewidth}{
    \begin{tabular}{@{}lccccc@{}}
        \toprule
        \textbf{Method} & \textbf{Easy} & \textbf{Medium} & \textbf{Hard} & \textbf{Extreme} & \textbf{Overall} \\
        \midrule
        Direct Answer & 52.49\% & 37.66\% & 17.55\% & 7.54\% & 17.47\% \\
        Standard CoT & 57.40\% & \textbf{58.45\%} & 24.46\% & 10.00\% & 28.46\% \\
        Reflective Reasoning & \textbf{62.00\%} & 47.46\% & \textbf{25.32\%} & \textbf{13.82\%} & \textbf{29.49\%} \\
        \midrule
        \textit{Gain (Reflect vs Direct)} & \textit{+9.51\%} & \textit{+9.80\%} & \textit{+7.77\%} & \textbf{\textit{+6.28\%}} & \textit{+12.02\%} \\
        \bottomrule
    \end{tabular}
    }
\end{table}

The results indicate a \textbf{Complexity-Dependent Scaling Law}:
\begin{itemize}
    \item \textbf{Low Complexity:} Standard CoT acts as a strong baseline, improving temporal grounding.
    \item \textbf{High Complexity (Extreme):} Reflective Reasoning achieves the highest gain (+6.28\% over Direct), confirming that self-correction is essential for long-context causal tracking.
\end{itemize}

\subsubsection{ Distractor Quantity Sensitivity ($K$)}

We investigated how model performance varies with the number of distractors $K$. We re-evaluated Qwen2.5-VL-72B on a subset of 500 random samples.

\begin{table}[h]
    \centering
    \small
    \caption{\textbf{Impact of Distractor Quantity ($K$).} Accuracy drops as the selection pool expands. The performance stabilization between $K=4$ and $K=5$ (only -1.9\% drop) identifies $K=4$ as the optimal saturation point for evaluation efficiency.}
    \label{tab:k_sensitivity}
    \adjustbox{max width=\linewidth}{
    \begin{tabular}{@{}lcccc@{}}
        \toprule
        \textbf{Metric} & \textbf{K=2} & \textbf{K=3} & \textbf{K=4 (Default)} & \textbf{K=5} \\
        \midrule
        Random Guess & 50.0\% & 33.3\% & 25.0\% & 20.0\% \\
        Model Accuracy & 68.4\% & 52.1\% & \textbf{42.3\%} & 41.5\% \\
        \midrule
        \textit{Relative Drop} & - & -23.8\% & \textbf{-18.8\%} & -1.9\% \\
        \bottomrule
    \end{tabular}
    }
\end{table}

\subsubsection{Prompt Robustness Test}

To ensure MM-CoT measures reasoning ability rather than prompt compliance, we tested three prompt variations on GPT-5. As shown in Table~\ref{tab:prompt_robustness}, the performance variance is minimal, confirming robustness.

\begin{table}[h]
    \centering
    \small
    \caption{\textbf{Prompt Robustness Analysis.} Comparison of GPT-5 performance under different instruction constraints.}
    \label{tab:prompt_robustness}
    \adjustbox{max width=\linewidth}{
    \begin{tabular}{@{}llc@{}}
        \toprule
        \textbf{Prompt Variation} & \textbf{Description} & \textbf{Accuracy} \\
        \midrule
        \textbf{Standard (Default)} & Full definition of A/B/C structure & \textbf{19.47\%} \\
        \textbf{Minimal} & ``Select the valid chain'' (No structure def.) & 18.21\% \\
        \textbf{Noisy} & Standard prompt w/ 2 typos per sentence & 19.10\% \\
        \midrule
        \textit{Variance} & Maximum Deviation & $\pm 1.26\%$ \\
        \bottomrule
    \end{tabular}
    }
\end{table}

\subsection{ Human Performance Study Details}

The ``Human'' performance column in Main Paper Table 1 serves as the theoretical upper bound. Table~\ref{tab:human_study} details the quantitative metrics from our study involving 20 graduate participants evaluating 200 stratified samples.

Participants achieved high agreement ($\kappa=0.89$), verifying that MM-CoT instances have objective ground truths. The primary error source for humans was missing fleeting details in \textit{Extreme} length videos, rather than logical reasoning failures.


\end{document}